\pdfoutput=1

\documentclass[a4paper,11pt,twoside]{article}
\usepackage[left=2.5cm,right=2cm,top=2cm,bottom=2cm]{geometry}
\usepackage[english]{babel}
\usepackage[utf8x]{inputenc}
\usepackage{graphicx}
\usepackage{hyperref}
\usepackage{placeins}
\usepackage{float}
\usepackage{listings}
\usepackage{subfigure}
\usepackage{caption}
\usepackage{setspace}
\usepackage[titletoc,toc,title]{appendix}
\hypersetup{colorlinks=true,citecolor=black,filecolor=black,linkcolor=black,urlcolor=black}

\usepackage[numbers,square]{natbib}

\begin{document}

	\title{
	\vspace{2cm}
	\huge Imperial College London \\
	\vspace{0.3cm}
	\large Department of Computing \\
    \vspace{5cm}
    \LARGE Giraffe: Using Deep Reinforcement Learning to Play Chess \\
    \vspace{1cm}
    \large by\\
    \vspace{1cm}
    \large {\href{mailto:m@matthewlai.ca}{Matthew Lai}} \\
    \vspace{5cm}
    \large Submitted in partial fulfilment of the requirements for the MSc Degree in \\ Advanced Computing of Imperial College London \\
    \vspace{1cm}
    \large September 2015 \\
	}
	
	\date{}

	\maketitle
	\setlength{\parindent}{0pt}

    \thispagestyle{empty}

\newpage
\begin{abstract}
\normalsize
\setlength{\parskip}{0.8em}

This report presents Giraffe, a chess engine that uses self-play to discover all its domain-specific knowledge, with minimal hand-crafted knowledge given by the programmer. Unlike previous attempts using machine learning only to perform parameter-tuning on hand-crafted evaluation functions, Giraffe's learning system also performs automatic feature extraction and pattern recognition. The trained evaluation function performs comparably to the evaluation functions of state-of-the-art chess engines - all of which containing thousands of lines of carefully hand-crafted pattern recognizers, tuned over many years by both computer chess experts and human chess masters. Giraffe is the most successful attempt thus far at using end-to-end machine learning to play chess.

We also investigated the possibility of using probability thresholds instead of depth to shape search trees. Depth-based searches form the backbone of virtually all chess engines in existence today, and is an algorithm that has become well-established over the past half century. Preliminary comparisons between a basic implementation of probability-based search and a basic implementation of depth-based search showed that our new probability-based approach performs moderately better than the established approach. There are also evidences suggesting that many successful ad-hoc add-ons to depth-based searches are generalized by switching to a probability-based search. We believe the probability-based search to be a more fundamentally correct way to perform minimax.

Finally, we designed another machine learning system to shape search trees within the probability-based search framework. Given any position, this system estimates the probability of each of the moves being the best move without looking ahead. The system is highly effective - the actual best move is within the top 3 ranked moves 70\% of the time, out of an average of approximately 35 legal moves from each position. This also resulted in a significant increase in playing strength.

With the move evaluator guiding a probability-based search using the learned evaluator, Giraffe plays at approximately the level of an FIDE International Master (top 2.2\% of tournament chess players with an official rating)\footnote{Fédération Internationale des Échecs, or the World Chess Federation, is the international organisation that governs all major international chess competitions.}\footnote{Please see Appendix~\ref{app:elo} for a description of the Elo rating system.}.

\end{abstract}

\newpage
\renewcommand{\abstractname}{Acknowledgements}
\begin{abstract}
\normalsize
\setlength{\parskip}{0.8em}

This project would not have been possible without the guidance, encouragement, insights, and inspiration from my supervisor Professor Duncan Gillies.

I would also like to thank the computer chess community for their support. In particular - Gerd Isenberg, for maintaining and writing most of the Chess Programming Wiki, the most comprehensive reference for everything to do with computer chess, and Graham Banks, for testing Giraffe against a wide variety of opponents.

I am also thankful for the hundreds of people who played thousands of games against Giraffe on the Internet Chess Club, including Grandmaster Tahir Vakhidov, International Master Alejandro Bofill Mas, IM Alejandro Montalvo, IM Alex Domont, IM Ulrich Schulze, IM William Hartston, FIDE Master 'renium', 'solsjenitsyn', 'arnav311004', 'wmm72', and 'DoctorNick'. Observing their games on ICC allowed me to discover many potential improvements for Giraffe.

This project would have been much less successful without the Imperial College High Performance Computing service, which provided the immense computing power required for this project.

And my family and friends, for all the fish.

\end{abstract}
	
	\newpage
	\tableofcontents
	\newpage
	\listoftables
	\newpage
	\listoffigures
	\newpage

\section{Introduction} 
\label{sec:introduction}

\subsection{Computers and Chess} 
\label{sec:computers_and_chess}
Chess is a game that requires so much creativity and sophisticated reasoning that it was once thought of as something no computers will ever be able to do. It was frequently listed alongside activities like poetry writing and painting, as examples of tasks that can only be performed by humans. While writing poetry has remained very difficult for computers to this day, we have had much more success building chess-playing computers.

In 1997, IBM's Deep Blue defeated the reigning World Chess Champion, Garry Kasparov, under standard tournament rules, for the first time in the history of chess. In the ensuing two decades, both computer hardware and AI research advanced the state-of-art chess-playing computers to the point where even the best humans today have no realistic chance of defeating a modern chess engine running on a smartphone.

However, it is interesting to note that the way computers play chess is very different from how humans play. While both humans and computers search ahead to predict how the game will go on, humans are much more selective in which branches of the game tree to explore. Computers, on the other hand, rely on brute force to explore as many continuations as possible, even ones that will be immediately thrown out by any skilled human. In a sense, the way humans play chess is much more computationally efficient - using Garry Kasparov vs Deep Blue as an example, Kasparov could not have been searching more than 3-5 positions per second, while Deep Blue, a supercomputer with 480 custom "chess processors", searched about 200 million positions per second \cite{hsu1995deep}\cite{hsu2002behind} to play at approximately equal strength (Deep Blue won the 6-game match with 2 wins, 3 draws, and 1 loss).

How can a human searching 3-5 positions per second be as strong as a computer searching 200 million positions per second? And is it possible to build even stronger chess computers than what we have today, by making them more computationally efficient? Those are the questions this project investigates.

There have been many attempts at making chess engines more selective without overlooking important continuations, but it has proven to be a very difficult problem. It is difficult to come up with reliable rules to decide which branches to explore further and which branches to prune. Humans do this by applying very complex rules with many exceptions. People learn those rules through playing and observing thousands of games, and no one has been able to convert that knowledge into concrete rules for computers that work in all situations.

Instead of the conventional method of teaching computers how to play chess by giving them hardcoded rules, this project is an attempt to use machine learning to figure out how to play chess through self-play, and have them derive their own rules from the games.

Using multiple deep artificial neural networks trained in a temporal-difference reinforcement learning framework, we use machine learning to assist the engine in making decisions in a few places -
\begin{itemize}
\item Statically evaluating positions - estimating how good a position is without looking further
\item Deciding which branches are most "interesting" in any given position, and should be searched further, as well as which branches to discard
\item Ordering moves - determining which moves to search before others, which significantly affects efficiency of searches
\end{itemize}

Using artificial neural networks as a substitute for "intuition", we hope to create a machine that can play chess more efficiently.

\section{Background and Related Work} 
\label{sec:background}

Thanks to the complex yet well-defined nature of the game, chess is an extensively-studied problem in artificial intelligence. Thousands of papers have been written on all aspects of computer chess since the 1950s, and at least hundreds of chess engines have been written by researchers, professionals, and amateurs over the same years \cite{ccrl4040}.

The chess problem is as follows - given a chess position, we want to find a move, out of all legal moves for the position, that maximizes our chance of winning the game. This means we also have to have a model for the opponent, for how she chooses her moves.

In high level chess, we usually model the opponent to be the same as ourselves - that is, we assume they select moves just as we do, and never make mistakes. This assumption may not be valid in games against much weaker players, where, for example, we may have two possible moves A and B, where move A results in a better position for us unless the opponent sees and plays an extremely complicated reply (that we saw but believe the opponent will overlook due to relative incompetence). It is risky to count on opponents making mistakes (or what seem to be mistakes from our perspective), so almost all chess engines implement the conservative model that assumes perfect opponents, to the limit of the engine's ability. From a game theory perspective, this strategy leads to Nash equilibrium - if one player adopts the strategy, the opponent can do no better than adopting the same strategy.

There is an alternative formulation of the chess problem that is more often used in practice. This formulation is as follows - given a chess position, we want to assign a score to it that corresponds to the chance of winning for the side to move. The only requirement for this score is that it should be monotonically increasing with respect to chance of winning. The exact relationship between score and winning probability is usually not known and not important.

Scores are usually centered around 0 (50\% chance of winning) and have odd symmetry for computational reasons. In that case, since chess is a zero-sum game, the score for one side is simply the negated score for the opposite side, for any given position.

It is easy to see that if we have a solution to the second problem, we can solve the first problem by simply selecting the move that results in the position with the lowest score (lowest chance of opponent winning). In this paper, we will mostly work with the second formulation.

\subsection{Conventional Chess Engines} 
\label{sec:conventional_engines}

Although they differ in implementation, almost all chess engines in existence today (and all of the top contenders) implement largely the same algorithms. They are all based on the idea of the fixed-depth minimax algorithm first developed by John von Neumann in 1928 \cite{neumann1928theorie}, and adapted for the problem of chess by Claude E. Shannon in 1950 \cite{shannon1950xxii}.

\subsubsection{Minimax} 
\label{sec:minimax}

The minimax algorithm is a simple recursive algorithm to score a position based on the assumption that the opponent thinks like we do, and also wants to win the game.

In its simplest form -

\begin{lstlisting}[caption=Simple minimax, label=simple_minimax, mathescape]
function minimax(position)
{
    if position is won for side to move:
        return 1
    else if position is won for the opponent:
        return -1
    else if position is drawn:
        return 0
        
    bestScore = -$\infty$
                
    for each possible move mv:
        subScore = -minimax(position.apply(mv))
        if subScore > bestScore:
            bestScore = subScore
                
    return bestScore
}
\end{lstlisting}

This algorithm works in theory, and also in practice for simpler games like tic-tac-toe (game tree size of at most $9!$ or $362880$). However, as it searches the entire game tree, it is impractical for games like chess, where the average branching factor is approximately 35, with an average game length of about 80 plies. The search tree size for chess is estimated to be about $10^{123}$ \cite{allis1994searching} ($10^{46}$ without repetitions \cite{chinchalkar1996upper}) which is more than one hundred orders of magnitudes higher than what is computationally feasible using modern computers. Therefore, a chess engine must decide which parts of the game tree to explore.

The most common approach is a fixed-depth search, where we artificially limit how far ahead we will look, and when we are at the end of the sub-tree we want to search, we call a static evaluation function that assigns a score to the position by analyzing it statically (without looking ahead). A very simple evaluation function is to simply add up pieces of both sides, each multiplied by a constant (eg. $Q=9$, $R=5$, $B=3$, $N=3$, $P=1$). The evaluation function is very important, and is one of the major areas of investigation for this project.

\begin{lstlisting}[caption=Depth-limited minimax, label=dl_minimax, mathescape]
function minimax(position, depth)
{
    if position is won for the moving side:
        return 1
    else if position is won for the non-moving side:
        return -1
    else if position is drawn:
        return 0
        
    if depth == 0:
        return evaluate(position)
        
    bestScore = -$\infty$
                
    for each possible move mv:
        subScore = -minimax(position.apply(mv), depth - 1)
        if subScore > bestScore:
            bestScore = subScore
                
    return bestScore
}
\end{lstlisting}

There are 3 changes -
\begin{itemize}
\item minimax() gets an additional parameter, depth, representing how deep we want to search this position
\item If we are at depth $0$, call evaluate and return the result
\item If we are at depth $> 0$, recursively call minimax as before, and pass in $depth - 1$ as the depth for the subtree
\end{itemize}

The search will now terminate in reasonable time, given a reasonable depth limit, but there is a very serious problem - the horizon effect.

In the algorithm above, we cut off all branches at the same distance from the root (called the horizon). This is dangerous because, for example, if in one branch, the last move happens to be QxP (queen captures pawn), we may score that position as a pretty good position, since we just won a pawn. However, what we don't see is that the pawn is defended by another pawn, and the opponent will take our queen the next move. This is actually a very bad position to be in.

This problem cannot be solved by simply increasing the depth limit, because no matter how deep we are searching, there will always be a horizon. The solution is quiescent search. The idea is that once we get to $depth == 0$, instead of calling evaluate() and returning, we enter a special search mode that only expands certain types of moves, and only call evaluate() when we get to a "quiet" and relatively stable position.

What types of moves to include in q-search is a matter of debate. All engine authors agree that winning captures (capturing a higher valued piece) should be included. Some agree that other captures should also be included. Many believe queen promotions should also be included. Some believe checks and check evasions should also be included. Some believe that pawn moves to the 7th rank (1 rank before promotion) should also be included.

There is a tradeoff to be made here - if we include too many moves, q-searches become too large, and we won't be able to search many plies in normal search. If we include too few moves, we may suffer from reduced forms of the horizon effect.

Q-searches are usually not depth-limited, and instead rely on the tree terminating. Trees will always terminate (usually reasonably quickly) since the number of possible captures are usually limited, and tend to decrease as captures are made.

\subsubsection{$\alpha$-$\beta$ Pruning}
\label{sec:alphabeta}

Without introducing any heuristics and with no loss of information, this algorithm can be optimized by introducing a "window" for each call to minimax(). The idea is that if the true score is below the lowerbound, that means the caller already has a better move, and therefore doesn't care about the exact value of this node (only that it is lower than the lowerbound). Conversely, if the true score is higher than the upperbound, that means the caller also doesn't care about the exact value, just the fact that it is higher than the upperbound. It may seem counter-intuitive at first to have an upperbound, but the reason for that is because chess is zero-sum, and upperbound and lowerbound switch places for each ply as we search deeper.

This optimization is called $\alpha$-$\beta$ pruning \cite{newell1976computer}, where $\alpha$ and $\beta$ are the lower and upper bounds. Since $\alpha$ and $\beta$ are horrible choices of variable names, we will refer to them as lowerbound and upperbound in the rest of this paper, but still refer to the algorithm as $\alpha$-$\beta$ for consistency with existing literature.

A detailed analysis of $\alpha$-$\beta$ is omitted here for brevity, but there is one very significant implication - that the order moves are searched is extremely important.

In standard minimax, move ordering (the order in which we explore nodes) is irrelevant because all nodes have to be visited exactly once. With $\alpha$-$\beta$ pruning, we only explore nodes that can potentially be "useful", and we can stop searching a node as soon as we prove that the result will be outside the window. That means, if we always expand the best nodes first, we will not have to examine as many nodes as if we had sub-optimal ordering.

In the worst case (opposite of optimal ordering), $\alpha$-$\beta$ degenerates into minimax. On the other hand, it has been proven that in the best case, where best moves are always searched first, $\alpha$-$\beta$ reduces effective branching factor to the square root of the original value, which would allow the search to go about twice as far in the same amount of time \cite{russell1995modern}.

It is obviously impossible to always have optimal ordering - since if there is a way to generate an optimal ordering for any given position, there would be no need to search at all. However, by ensuring that move ordering is close to optimal in most cases using heuristics, we can make the search more efficient.

In conventional chess engines, there are a few common heuristics to improve move ordering -
\begin{itemize}
\item If we had previously searched the same position, the move that ended up being the best is probably still the best (even if the previous search was done to a shallower depth)
\item If a move proved to be good in a sibling node, there is a good chance it will be good for the node under consideration as well
\item Capturing high-valued pieces with low-valued pieces is usually good, and capturing defended low-valued pieces with high-valued pieces is usually bad
\item Queen promotions are usually good
\end{itemize}

In this project, we will investigate using machine learning to assist the engine in ordering moves.

Depth-limited minimax() with $\alpha$-$\beta$ pruning and q-search form the backbone of virtually all existing chess engines. There are many further optimizations that can be done and have been done to further improve the search. Those optimizations will not be covered in this paper because they are not directly relevant for this project. However, many of them have been implemented in the chess engine created for this project.

\subsubsection{Evaluation} 
\label{sec:evaluation}

As mentioned above in Section \ref{sec:minimax}, the evaluation function is a very important part of a chess engine, and almost all improvements in playing strength among the top engines nowadays come from improvements in their respective evaluation functions.

The job of the evaluation functions is to assign scores to positions statically (without looking ahead). Evaluation functions contain most of the domain-specific knowledge designed into chess engines.

In this project, we will develop an evaluation function based on a machine learning approach. However, before we do that, we will take a look at the evaluation function of Stockfish \cite{1_stockfishchess.org_2015}, an open source chess engine that is currently the strongest chess engine in the world. Examining an existing state-of-art evaluation function can help us design an effective feature representation for our machine learning effort.

Stockfish's evaluation function consists of 9 parts -
\begin{itemize}

\item \textbf{Material}: Each piece on the board gets a score for its existence. Synergetic effects are also taken into account (for example, having both bishops gives a bonus higher than two times the material value of a bishop), and polynomial regression is used to model more complex interactions.

\item \textbf{Piece-Square Tables}: Each piece gets a bonus or penalty depending on where they are, independent of where other pieces are. This evaluation term encourages the engine to advance pawns and develop knights and bishops for example.

\item \textbf{Pawn Structure}: The position is scanned for passed, isolated, opposed, backward, unsupported, and lever pawns. Bonuses and penalties are assigned to each feature. These features are all local, involving 2-3 adjacent pawns.

\item \textbf{Piece-specific Evaluation}: Each piece is evaluated individually, using piece-type-specific features. For example, bishops and knights get a bonus for being in a "pawn outpost", rooks get a bonus for being on an open file, semi-open file, or the same rank as enemy pawns.

\item \textbf{Mobility}: Pieces get bonuses for how many possible moves they have. In Stockfish's implementation, squares controlled by opponent pieces of lesser value are not counted. For example, for bishop mobility, squares controlled by enemy pawns are not included, and for queen mobility, squares controlled by bishops, knights, or rooks are not included. Each piece type has a different set of mobility bonuses.

\item \textbf{King Safety}: Bonuses and penalties are given depending on number and proximity of attackers, completeness of pawn shelter, and castling rights.

\item \textbf{Threat}: Undefended pieces are given penalties, defended pieces are given bonuses depending on piece type, and defender type.

\item \textbf{Space}: Bonuses are given for having "safe" empty squares on a player's side.

\item \textbf{Draw-ish-ness}: Certain material combinations often result in draws, so in these cases, the evaluation is scaled toward 0.

\end{itemize}

It is quite a complicated function with a lot of hand-coded knowledge. Most engines don't have evaluation functions that are nearly as extensive, because it is difficult to tune such a high number of parameters by hand.

\subsection{Machine Learning in Zero-Sum Games} 
\label{sec:ml_zs}

Using machine learning to play zero-sum games (including chess) has been attempted numerous times. This section summarizes such attempts, and explains how they relate to this project.

\subsubsection{Mate Solvers}
\label{sec:mate_solvers}

There have been many attempts at using machine learning to solve chess endgames. Endgames are usually defined as positions with only a few pieces (usually less than 6 or so in total) left. These situations are much simpler than average chess positions, and relatively simple rules for optimal play can often be derived. Most attempts in this area use genetic programming \cite{hauptman2005gp} \cite{lassabe2006genetically}, and inductive learning \cite{o1982comparative}. Unfortunately, those approaches do not scale to much more complicated positions seen in midgames and openings, where no simple rules for optimal play exist.

\subsubsection{Learning Based on Hand-Selected Features} 
\label{sec:hand_selected_features}

In a typical chess evaluation function, there are hundreds of parameters that must be tuned together. As it is very time-consuming to do the tuning by hand, some attempted to either tune those parameters automatically using machine learning, or use the hard-coded features as input to machine learning systems to generate final evaluation scores. NeuroChess \cite{baxter1998experiments} is one such attempt. The Falcon project \cite{david2008genetic} is an attempt to tune evaluation parameters using genetic programming, with the help of a mentor, which has a much more complicated evaluation function. While it is possible to achieve good performance using this method, the fact that features are hand-picked means the systems are limited by the creativity of their designers (and the performance of the mentor in the case of Falcon). A more recent attempt is the Meep project by Veness et al. \cite{veness2009bootstrapping}, whose evaluation function is a linear combination of hand-designed features, and where the weights are trained using reinforcement learning.

\subsubsection{Temporal-Difference Learning} 
\label{sec:td_learn}

All machine learning approaches require a way to evaluate the performance of models. While results from gameplay is the most direct way to measure the performance of models, it is often not practical due to the very large number of games required to get statistically significant results for small changes \cite{shawulpaired}. It is computationally inefficient to perform all the calculations required to play a game, only to get a result with less than 2 bits of information (won/drawn/loss).

The most successful approach thus far is temporal-difference reinforcement learning, where the system is not tested on its ability to predict the outcome of the game from a position, but on its ability to predict its own evaluation in the near future, and therefore achieving temporal consistency. As long as some positions have fixed score (eg. won, lost, or drawn positions), models with higher predictive power would have better temporal consistency (this is not true if no position has fixed score, since the model can always produce a constant value for example, and achieve temporal consistency that way \cite{thrun1995learning}). This approach has several advantages over using game results - the most important one being that evaluations used to make moves in the same game can receive different error signals. In chess, it is possible for a game to be lost by one blunder, even if all other moves are optimal. In a system based on game results, all the moves in the game (even the good ones) will receive the high error signal, whereas in temporal-difference learning, only the blunder (which creates a temporal inconsistency) will get a high error signal. This significantly improves the signal-to-noise ratio in error signals.

The most famous study in using temporal-difference reinforcement learning to play zero-sum games is probably Tesauro's backgammon program TD-Gammon \cite{tesauro1995temporal} from 1995. The system was trained from random initialization using temporal-difference reinforcement learning through self-play. TD-Gammon attained master-level performance after playing 1.5 million training games, greatly surpassing all previous programs. TD-Gammon uses neural networks with raw board input as well as hand-designed features for board representation. There is no search involved. From any given position, TD-Gammon simply picks the move that results in a position with the best evaluation score.

The same approach was used on Go by Schraudolph et al. \cite{schraudolph1994temporal}, but they were only able to achieve "a low playing level" \cite{schraudolph1994temporal}, due to the complexity of the game, and the necessity to look ahead.

A similar approach was used by Thrun in his NeuroChess program \cite{thrun1995learning}, which, when used with GNU Chess's search routine, wins approximately 13\% of all games against GNU Chess. The author attributed the disappointing performance to inefficiency in implementation, as neural network evaluation, as implemented in NeuroChess, takes "two orders of magnitudes longer than evaluating an optimized linear evaluation function (like that of GNU-Chess)" \cite{thrun1995learning}.

In 1999, Baxter et al. adapted the temporal-difference learning algorithm to be used in conjunction with minimax searches \cite{baxter1999tdleaf}. They called it TDLeaf($\lambda$). The idea is that since in minimax, the final score returned is always the evaluation score of the leaf position of the principal variation, it is more efficient to perform T-$\Delta$ on evaluation of the leaf nodes, instead of the root nodes. They enjoyed enormous success using this approach, with their KnightCap program achieving FIDE International Master strength after about 1000 games played on an internet chess server. It was also able to easily defeat other top chess engines when they are constrained to the same depths reached by KnightCap, proving that KnightCap has a superior evaluation function. However, due to its much slower evaluation function (5872 hand-designed features), KnightCap was weaker than top engines of the time in tournament conditions. The authors suggested that it is very difficult to compensate for lack of depth using superior evaluation, since tactics make chess highly non-smooth in feature-space, which makes machine learning very difficult. Although not explored in their study, they also postulate that without significantly simplifying (and speeding up) the evaluation function, the best way to solve tactical myopia is by finding "an effective algorithm for [learning] to search [selectively]," which would have "potential for far greater improvement" \cite{baxter1999tdleaf}. This is another one of the main topics of investigation for this project.

\subsection{Deep Learning} 
\label{sec:deep_learning}

To go beyond weight tuning with hand-designed features and actually have a learned system perform feature extraction, we need a powerful and highly non-linear universal function approximator which can be tuned to approximate complex functions like the evaluation function in chess.

Most machine learning models perform well only for approximating functions that are first or second order. They are not capable of building up hierarchies of knowledge to model very complex high order features. As a result, in order to approximate complex functions where the input-output mapping is inherently hierarchical, the models would have to be made extremely large. Extremely large models not only require high computational power, but they also require very large training sets to prevent overfitting, due to the high degrees of freedom in the model. Unfortunately, computational power and availability of training examples are usually limiting factors in practical machine learning applications.

As an example of problems requiring hierarchical feature extraction, consider the problem of classifying whether an image contains a car or not, given pixel intensities of the image.

To perform the classification directly from pixels, the system would have to essentially evolve to memorize images of cars, and test input images for similarity with those images. Besides the problems mentioned above, these models are also unlikely to generalize well - given only red, blue, and yellow cars in the training set, it's unlikely to be able to correctly identify green cars.

The most efficient and accurate way to classify those images is with a hierarchical approach, where the first layer would identify low level features like gradients, corners, and edges. The second layer can then use the output of the first layer to identify shapes. Finally, the location of the shapes can be used to classify objects in the image. This is what humans naturally do, and it's something that machine learning systems struggled with until recently.

A new class of algorithms capable of performing hierarchical knowledge extraction were popularized in 2007 by Hinton et al.\cite{hinton2007learning}. While there are a few models that can be extended to perform deep learning, the vast majority of current applications use artificial neural networks.

In the past few years, deep learning has refreshed the state-of-the-art results on a wide range of problems, often greatly surpassing the previous results. For example, in 2012, Krizhevsky et al. trained a deep network for the popular ImageNet competition (classifying 1.2 million images into 1000 different classes), achieving a top-5 error rate of 15.3\%, compared to the previous best of 26.2\% \cite{krizhevsky2012imagenet}. It has also been used for handwriting recognition, where deep networks are approaching human performance at 0.3\% error \cite{krizhevsky2012imagenet}.

As another example, Google's DeepMind recently published their results on training deep networks to play seven different arcade games, given only pixel intensities as input, and increasing the score displayed as the objective. The networks they trained were able to surpass the performance of previous game-specific AIs on six of the seven games, and exceeded human expert performance on three of them \cite{mnih2013playing}.

Besides computer vision and control tasks, deep learning has also been successfully applied to natural language processing, sentiment analysis, audio analysis, and other fields where knowledge is inherently hierarchical.

In this project, we apply deep learning to chess. We use deep networks to evaluate positions, decide which branches to search, and order moves.

\subsection{Project Overview and Contributions} 
\label{sec:overview}

This project aims to create a new chess engine based on machine learning, using both algorithms that have already been proven to be successful in past attempts such as TDLeaf($\lambda$), as well as new algorithms developed for this project.

The first part of the project is essentially a modern iteration of the TDLeaf($\lambda$) experiment \cite{baxter1999tdleaf} on learning new evaluation functions, taking advantage of discoveries and advances in neural networks in the past 15 years, as well as the much higher computational power available today. Our aim is to create a system that can perform high level feature extraction automatically, in order to not be constrained by human creativity - something that was not practical in the 1990s due to computational power constraints.

In the second part of the project, we develop a novel probability-based search framework to enable highly selective searches using neural networks to provide guidance on the amount of time to spend on each subtree from any given position. This approach is highly speculative, and challenges decades-worth of established research on using distance-from-root as the primary factor in deciding which subtrees to explore in minimax searches. It is already well-established that strategic extensions, reductions, and pruning have huge benefits on the performance of chess engines \cite{hoki2012efficiency} \cite{heinz2000adaptive}, however, existing heuristics for making those decisions are mechanical and quite crude, and must be done very conservatively to not miss tactical details. It is hoped that this project will bring the two fields - machine learning and game theory, together in a more intimate way than has ever been done.

\section{Supporting Work} 
\label{sec:supporting_work}

A chess engine, Giraffe, has been written to be used as the basis for modifications. Most of the popular techniques in computer chess have been implemented in this chess engine. With a basic linear evaluation function that takes into account material, piece-square tables, mobility, and some king safety, it plays at slightly below FIDE Candidate Master strength.

The search routine is also fairly standard, though aggressive heuristic pruning techniques have been intentionally omitted, because we want all the heuristic decisions to be made by learned systems.

An implementation of neural networks has also been written. A decision was made to write one from scratch instead of using an existing framework since it needs to be extensively customized for our application for features such as restricted connectivity, and using an existing framework would not save a lot of work, and would involve making extensive modifications to an unfamiliar codebase.

The implementation is done in C++, and is fully vectorized using the Eigen high performance linear algebra library \cite{eigenweb}, and multi-threaded to achieve performance comparable to other optimized machine learning libraries. A few state-of-art optimization algorithms have also been implemented - mini-batch stochastic gradient descent with Nesterov's accelerated momentum \cite{sutskever2013importance}, the AdaGrad adaptive subgradient method \cite{duchi2011adaptive}, and an improved version of the AdaGrad method known as AdaDelta \cite{zeiler2012adadelta}. Both AdaGrad and AdaDelta maintain individual learning rates for each parameter, so that for example, nodes that are rarely activated can use a higher learning rate to take full advantage of the few activations, while nodes that are often activated can use lower learning rate to converge to better minimums. This is very important for chess, where many patterns are rarely present. Experiments have agreed with the results presented in the AdaGrad and AdaDelta papers, and we do get the fastest convergence and highest quality minimums with AdaDelta.

The following sections will describe modifications to the engine to replace parts of it with learned systems.

\section{Neural Network-Based Evaluation}
\label{sec:nn_eval}

The first problem we tackle is the evaluation function, because other systems (eg. move ordering) require us to already have a good evaluation function. As mentioned above in Section~\ref{sec:evaluation}, the job of the evaluation function is to take a position as an input, and estimate the probability of winning for the players, without searching forward. This function is called at the leaf nodes of searches, when we have to terminate the search due to time constraint.

\subsection{Feature Representation}
\label{sec:representation}

Feature representation is one of the most critical part of the project. The features should be of low enough level and be general enough that most of the chess knowledge is still discovered through learning. On the other hand, a well designed board representation can significantly improve the performance of the system, by making useful features easier to learn. The goal of the representation is not to be as succinct as possible, but to be relatively untangled, and smooth in feature space. For example, while in principle, neural nets can learn the rules of chess and figure out that the influence of sliding pieces ends at the first piece along each direction, it is much more efficient to give it this information as part of the board representation.

For neural networks to work well, the feature representation needs to be relatively smooth in how the input space is mapped to the output space. Positions that are close together in the feature space should, as much as possible, have similar evaluations.

This means many intuitive and naive representations are not likely to work well. For example, many previous attempts at using neural networks for chess represent positions as bitmaps, where each of the 64 squares is represented using 12 binary features indicating whether a piece of each of the 12 piece types exist on the square or not. This is unlikely to work well, because in this 768-dimensional space, positions close together do not necessarily have similar evaluation. As an illustration, see Figure~\ref{fig:feat_rep_ex} below. In this case, the position on the left should have similar evaluation to the position in the middle, because the bishops are in the same region of the board. The position on the right should have a very different evaluation, because the bishop is somewhere else entirely.

\FloatBarrier
\begin{figure}[h]
\hfill
\subfigure[Similar Position]{\includegraphics[scale=0.5]{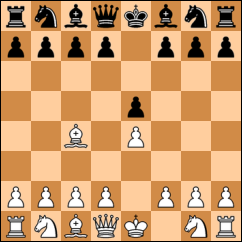}}
\hfill
\subfigure[Original]{\includegraphics[scale=0.5]{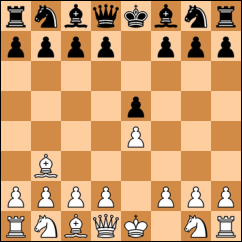}}
\hfill
\subfigure[Different Position]{\includegraphics[scale=0.5]{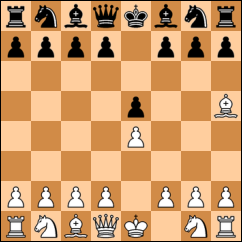}}
\hfill
\caption{Example to illustrate effects of different feature representations}
\label{fig:feat_rep_ex}
\end{figure}

In the bitmap feature representation, the 3 positions all have the same distance to each other in the feature space. That means the network will have a hard time generalizing the advantages of having pieces on specific locations to general regions on the board.

A much better representation is to encode the position as a list of pieces and their coordinates. This way, positions that are close together in the feature space would usually have similar evaluation. Since the feature vector needs to be the same length regardless of how many pieces are on the board, we use a slot system. The first two slots are reserved for the kings, followed by two slots for the queens, then four slots for the knights, four slots for the bishops, four slots for the rooks, and finally sixteen slots for the pawns. In addition to coordinates, each slot also encodes presence or absence of the piece, as well as some additional information about each piece such as whether its defended or not, and how far can it move along each direction (for sliding pieces). In case of promotions it is possible to have more than the starting number of some of the pieces for some piece types. In those very rare cases, the extra pieces are not encoded in the slots system, though piece counts are encoded separately, so they will be scored for their existence (but not position).

We evaluated board representations by training neural networks to predict the output of Stockfish's evaluation function in a supervised fashion, given 5 million positions as input, in the board representation under evaluation. The positions are uniformly randomly sampled from databases of high quality games, either between human grandmasters or between computers. They include games in all phases and situations, and they are all labeled by Stockfish.

It is hoped that if we can find a board representation that allows a network to efficiently learn Stockfish's evaluation function, it will be flexible enough to model at least the kind of chess knowledge we have right now, and hopefully unknown knowledge in similar format in the future. The challenge is to achieve that while keeping the board representation as general as possible, to not limit its ability to learn new things.

In the end, we arrived at a feature representation consisting of the following parts -
\begin{itemize}
\item \textbf{Side to Move} - White or black
\item \textbf{Castling Rights} - Presence or absence of each of the 4 castling rights (white long castle, white short castle, black long castle, black short castle)
\item \textbf{Material Configuration} - Number of each type of pieces
\item \textbf{Piece Lists} - Existence and coordinates of each (potential) piece. There is 1 slot per side for kings, 1 slot for queens, 2 slots for rooks/bishops/knights, and 8 slots for pawns. Each slot has normalized x coordinate, normalized y coordinate, whether piece is present or absent, and the lowest valued attacker and defender of the piece.  Pieces are assigned slots based on their position when possible. For example, if only a, b, g, h pawns are present, they will occupy the 1st, 2nd, 7th, and 8th slots respectively.
\item \textbf{Sliding Pieces Mobility} - For each sliding piece slot, we encode how far they can go in each direction (before being stopped by either the board's edge, or a piece). We also tried the same for non-sliding pieces (kings, knights, pawns), and saw no improvement, showing that the neural network is powerful enough to learn their movement patterns without much trouble.
\item \textbf{Attack and Defend Maps} - The lowest-valued attacker and defender of each square.
\end{itemize}

There are 363 features in total.

The piece lists make up the bulk of the feature representation. They allow the network to efficiently learn things like centralisation of knights and bishops, rooks on opponent's pawn rank, centralisation of the king in end games, and advancement of pawns in general. They also allow the network to learn the importance of mobility for different pieces in different phases of the game, as well as some concepts of king safety based on distances between different types of attacking pieces and the opponent's king. As it turned out, these features are by far the most important features, because many important concepts can be represented as linear combinations of pieces' coordinates, or linear combinations of pieces' coordinates with thresholds. In fact, the system performs at a reasonable level with just these features.

The side to move flags is added so the network can learn the concept of tempo in chess - the fact that all else being equal, it's better to be the moving side, because in almost all situations, there is an available move that is better than passing (making no move). This is not always true. In particular, there is a class of positions known as Zugzwang, where all legal moves worsen the situation for the moving side, but the moving side still has to make a move because passing is illegal in chess. The value of a tempo also depends on game phase, material configurations, as well as pawn structure. These are all left in the hands of the neural network to discover.

The material configuration section is almost entirely redundant, since the same information can be derived from the existence flags in the piece slots. However, they are still useful in cases of promotions, where we may run out of piece slots. With piece count features, at least they can be given a bonus for their existence. Also, having these numbers allow the network to more easily discover concepts like piece synergies, and how material combinations can affect the importance of other features - for example, moving the king towards the centre is often good in the end game, but very dangerous in the opening and middle game, where the opponent still has many pieces on the board.

The attack and defend maps allow the network to learn concepts related to board control. This is difficult for the network to derive from the piece lists, because piece lists are piece-centric and not square-centric. That is why these maps are provided even though the information is theoretically computable from other features.

\begin{figure}[ht!]
    \centering
    \includegraphics[scale=0.75]{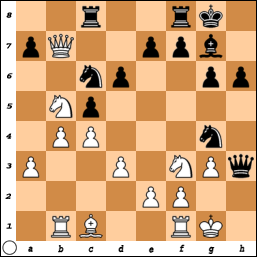}
    \caption{Feature representation example}
    \label{fig:feat_conv}
\end{figure}
\FloatBarrier

For example, the position in Figure~\ref{fig:feat_conv} above would be represented as follows (mobility, attack and defend maps omitted for brevity) in Table~\ref{tab:ex_pos_rep}:

\FloatBarrier
\begin{table}[ht!]
\centering
\begin{tabular}{|l|l|}
\hline
{\bf Feature}           & {\bf Value} \\ \hline
Side to Move            & White \\ \hline
White Long Castle       & No    \\ \hline
White Short Castle      & No    \\ \hline
Black Long Castle       & No    \\ \hline
Black Short Castle      & No    \\ \hline
White Queens            & 1     \\ \hline
White Rooks             & 2     \\ \hline
White Bishops           & 1     \\ \hline
White Knights           & 2     \\ \hline
White Pawns             & 7     \\ \hline
Black Queens            & 1     \\ \hline
Black Rooks             & 2     \\ \hline
Black Bishops           & 1     \\ \hline
Black Knights           & 2     \\ \hline
Black Pawns             & 7     \\ \hline
White Queen 1 Exists    & Yes   \\ \hline
White Queen 1 Position  & b7    \\ \hline
White Rook 1 Exists     & Yes   \\ \hline
White Rook 1 Position   & b1    \\ \hline
White Rook 2 Exists     & Yes   \\ \hline
White Rook 2 Position   & f1    \\ \hline
White Bishop 1 Exists   & Yes   \\ \hline
White Bishop 1 Position & c1    \\ \hline
White Bishop 2 Exists   & No    \\ \hline
White Bishop 2 Position & N/A   \\ \hline
...                     &       \\ \hline
\end{tabular}
\caption{Example position representation}
\label{tab:ex_pos_rep}
\end{table}

\subsection{Network Architecture}
\label{sec:network_arch}

The evaluator network is a 3-layer network (two hidden layers plus output layer), where all hidden nodes use Rectified Linear activation (ReLU) \cite{glorot2011deep}, which is a modern choice that is much more efficient than the traditional hyperbolic tangent and logistic activation functions, especially for networks with more than two hidden layers. To constrain the output between -1 and 1, the output node uses hyperbolic tangent activation.

\begin{figure}[ht!]
    \centering
    \includegraphics[scale=0.75]{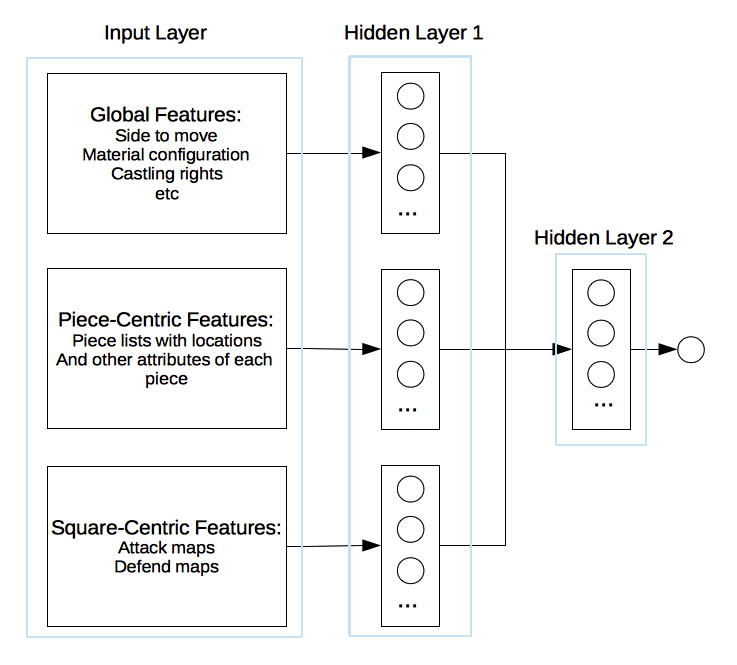}
    \caption{Network architecture}
    \label{fig:network_arch}
\end{figure}

When the feature representation includes features with different modalities, it is often beneficial to only mix them in higher, more abstract layers. This is because at very low levels, data from different modalities cannot be mixed usefully, and the extra connections offer no benefit (and in fact makes the model more susceptible to overfitting). In this case, there are three modalities - piece-centric, square-centric, and position-centric. In the first two layers, features from different modalities are kept separate. Only the final two layers are fully connected to capture interactions between high level concepts derived from features from the different modalities. Figure~\ref{fig:network_arch} illustrates the network architecture and the limited connectivity scheme, where an arrow means all nodes in the first group are connected to all nodes in the second group.

\FloatBarrier

\subsection{Training Set Generation}
\label{sec:training_set_gen}

Before we can start training the network, we need a way to generate a corpus of positions to train the system on. Generating a good corpus is not an easy task, because it needs to simultaneously satisfy a few potentially conflicting objectives -

\begin{itemize}
\item \textbf{Correct Distribution} - The positions should have approximately the same distribution as those that actually come up in game-play. For example, it doesn't make sense to train the system on positions with 3 queens per side, because those positions virtually never come up in actual games.
\item \textbf{Variety} - Although we want the positions to be representative of those that appear in actual games, we also want to have some variety on top of that, so that the system can learn to play in highly unequal positions for example. This is important because although they don't frequently appear in real games, they are encountered in internal nodes of searches all the time, and the engine must also know how to evaluate them properly.
\item \textbf{High Volume} - To train a highly flexible model, we need a large number of training positions. This requirement precludes any kind of human-based position generation (eg. online play).
\end{itemize}

Several methods have been used in previous attempts at applying machine learning to chess. For Falcon (an engine based on genetic programming), a 10,000-games database of grandmaster games is used, with 1 position drawn from each game \cite{david2008genetic}. A clever trick is applied to ensure variety - by changing every position to be white-to-move. This gives them more positions with material imbalances than would typically appear in grandmaster-level games \cite{david2008genetic}.

For KnightCap (an engine based on using TD-Leaf to tune the parameters of a hand-coded evaluation function), online play on the Free Internet Chess Server is used \cite{baxter1999tdleaf}. The engine played about 400 games for training. The authors also claimed that self-play is not possible due to limited variations in self-play (the engine would always play the same moves).

In this project, we used another approach that is a hybrid of existing approaches. First, we collected 5 million positions randomly from a database of computer games \cite{ccrl4040}. However, instead of using each position directly, we introduce imbalance and more unusual positions by randomly applying a legal move to each position before using it for training. These positions are then used as starting points for self-play.

This new approach satisfies all three criteria mentioned above - all the positions are slightly different from actual gameplay positions, there is a large variety of positions due to the random moves, and a total of about 175 million positions. Since we are performing reinforcement learning, these positions do not need to be labeled.

\subsection{Network Initialization}
\label{sec:eval_init}

Although TD-Leaf can train a system from random initialization in principle, for a complex problem like chess, random initialization would lead to extremely long training times. Therefore, we use a very simple evaluation function containing only very basic material knowledge to bootstrap the training process. The bootstrapping process only takes a few seconds, and puts the neural network at a point in the parameter space that is much closer to a good minimum.

It is also possible to bootstrap with a much more accurate function such as one of the top chess engine's evaluation, but as the goal of this project is to investigate how much chess knowledge can be learned entirely through self-discovery, this approach was not taken.

\subsection{TD-Leaf}
\label{sec:tdleaf}

To train the network, we need a way to generate error signals. It is possible to use another stronger chess engine as a mentor to generate ground-truth labels (as has been done for the Falcon project \cite{david2008genetic}), but for this project, we are aiming to create a system that can learn to play chess with no external help, and as little built-in chess knowledge as possible.

As mentioned in Section~\ref{sec:td_learn}, one popular way to do that is through temporal-difference reinforcement learning - trying to make the evaluation function a better predictor of its own evaluation later in time, and achieving temporal consistency. In this project, we use a modified version of the algorithm known as TD-Leaf($\lambda$), which allows faster learning by exploiting the fact that the local gradient of minimax is simply the gradient of the evaluation function at the leaf node of the search \cite{baxter1999tdleaf}.

In our implementation of TD-Leaf($\lambda$), for each training iteration, we randomly select 256 positions from the training set (with 1 random move applied to each), then have the engine play against itself for 12 moves. The results of the 12 searches are recorded. After the 12 moves have been played, we look at how the position score changes over the 12 moves, and compute an error for the starting position by adding up all the changes, weighted by how far they are from the starting position. Errors are scaled by $\lambda^{m}$, where $\lambda$ is fixed to be 0.7, and $m$ is how many moves away the error is from the starting position.

The update rule for TD-Leaf($\lambda$) is as follows -

\[
w = w + \alpha \sum_{t=1}^{N-1} \nabla J(x_t, w) \bigg[ \sum_{j=t}^{N-1} \lambda^{j-t} d_t \bigg]
\]

where $w$ is the set of weights of the model, $\alpha$ is the learning rate (set to 1 in our case), $\nabla J(x_t, w)$ is the gradient of the model at each time point, and $d_t$ is the temporal differences from each time point $d_t$ to the next time point \cite{baxter1999tdleaf}.

\FloatBarrier

\begin{table}[ht]
\centering
\caption{TD-Leaf Example}
\label{tab:tdl_example}
\begin{tabular}{llll}
\hline
\multicolumn{1}{|l|}{{\bf Search Score}} & \multicolumn{1}{l|}{{\bf Score Change}} & \multicolumn{1}{l|}{{\bf Discount Factor}} & \multicolumn{1}{l|}{{\bf Contribution to Total Error}} \\ \hline
\multicolumn{1}{|l|}{10}                 & \multicolumn{1}{l|}{0}           & \multicolumn{1}{l|}{$0.7^0$}               & \multicolumn{1}{l|}{0}                                 \\ \hline
\multicolumn{1}{|l|}{20}                 & \multicolumn{1}{l|}{10}          & \multicolumn{1}{l|}{$0.7^1$}               & \multicolumn{1}{l|}{7}                                 \\ \hline
\multicolumn{1}{|l|}{20}                 & \multicolumn{1}{l|}{0}           & \multicolumn{1}{l|}{$0.7^2$}               & \multicolumn{1}{l|}{0}                                 \\ \hline
\multicolumn{1}{|l|}{20}                 & \multicolumn{1}{l|}{0}           & \multicolumn{1}{l|}{$0.7^3$}               & \multicolumn{1}{l|}{0}                                 \\ \hline
\multicolumn{1}{|l|}{-10}                & \multicolumn{1}{l|}{-30}         & \multicolumn{1}{l|}{$0.7^4$}               & \multicolumn{1}{l|}{-7.2}                              \\ \hline
\multicolumn{1}{|l|}{-10}                & \multicolumn{1}{l|}{0}           & \multicolumn{1}{l|}{$0.7^5$}               & \multicolumn{1}{l|}{0}                                 \\ \hline
\multicolumn{1}{|l|}{-10}                & \multicolumn{1}{l|}{0}           & \multicolumn{1}{l|}{$0.7^6$}               & \multicolumn{1}{l|}{0}                                 \\ \hline
\multicolumn{1}{|l|}{40}                 & \multicolumn{1}{l|}{50}          & \multicolumn{1}{l|}{$0.7^7$}               & \multicolumn{1}{l|}{4.12}                              \\ \hline
\multicolumn{1}{|l|}{40}                 & \multicolumn{1}{l|}{0}           & \multicolumn{1}{l|}{$0.7^8$}               & \multicolumn{1}{l|}{0}                                 \\ \hline
\multicolumn{1}{|l|}{40}                 & \multicolumn{1}{l|}{0}           & \multicolumn{1}{l|}{$0.7^9$}               & \multicolumn{1}{l|}{0}                                 \\ \hline
\multicolumn{1}{|l|}{40}                 & \multicolumn{1}{l|}{0}           & \multicolumn{1}{l|}{$0.7^{10}$}              & \multicolumn{1}{l|}{0}                                 \\ \hline
\multicolumn{1}{|l|}{40}                 & \multicolumn{1}{l|}{0}           & \multicolumn{1}{l|}{$0.7^{11}$}              & \multicolumn{1}{l|}{0}                                 \\ \hline
                                         &                                  & \multicolumn{1}{r}{Total Error:}           & 4.10                                                  
\end{tabular}
\end{table}

\FloatBarrier

In the example in Table~\ref{tab:tdl_example}, the evaluation changed three times over the 12 moves - after moves 1, 4, and 7. Each of the score changes contributed to the error for the starting position.

The rationale for the use of the discount factor is the credit assignment problem - when there is a temporal inconsistency (score change), we know that one of the earlier evaluations were not accurate. However, there is no way to know which move actually caused the inconsistency. Therefore, we make the assumption that the move immediately before the score change most likely caused it, followed by the move before that, etc. This allows the evaluation function to learn to model patterns with long term consequences, but prioritize patterns with shorter term consequences.

Once the total errors for 256 positions are obtained, standard back-propagation is done to derive the gradient of the $L1$ loss function with respect to connection weights and biases of the neural network. $L1$ loss is chosen over the more common $L2$ loss because TD-Leaf generates errors with many outliers, and they would completely dominate $L2$ loss.

The gradients are then used to train the network using stochastic gradient descent with AdaDelta update rules \cite{zeiler2012adadelta}. AdaDelta is a recently discovered optimization technique that performs better than conventional gradient descent algorithms. Many different optimizations were implemented for this project - SGD with momentum, SGD with Nesterov's accelerated gradient \cite{nesterov1983method}, and adaptive gradient (AdaGrad) \cite{duchi2011adaptive}. AdaDelta was found to have the highest convergence rate, and converges to the best minimums.

The primary difference between AdaDelta and older techniques (besides AdaGrad) is that it maintains separate learning rates for each weight, and the learning rates are changed after each iteration based on the direction of the gradients. It allows weights for neurons that are rarely activated to retain higher learning rates to make best use of limited activations, while allowing weights for neurons that are often activated to decrease over time so that they can converge to better minimums \cite{zeiler2012adadelta}. This is similar in idea to the much older resilient back-propagation (RPROP) algorithm \cite{riedmiller1992rprop}, which does not work in a stochastic setting due to gradients frequently changing signs.

\subsection{Results and Evaluation}
\label{sec:eval_results}

To test the positional understanding of the evaluator, the engine is run through the Strategic Test Suite \cite{sts}. The Strategic Test Suite is a collection of 1500 positions, split into 15 themes of 100 positions each. Each group of 100 positions tests the engine's understanding on one strategic idea. For example, one theme tests the understanding of control of open files, another tests the understanding of how bishop and knight's values change relative to each other in different situations, and yet another tests the understanding of centre control. Unlike most other test suites, the STS positions do not have deep tactical lines for the engine to find, making it especially suitable for evaluating positional understanding of an engine.

Each of the positions in the test suite have a list of solution moves with associated scores. The best solutions always have a score of 10, and other solutions have scores between 0 and 10. At the end of a test run, the scores for the engine's moves for each of the position are added up to a score out of 15000.

Positions from the test suite have never been used in any stage of training. It is a completely independent test set.

\begin{figure}[ht!]
    \centering
    \includegraphics[scale=0.6]{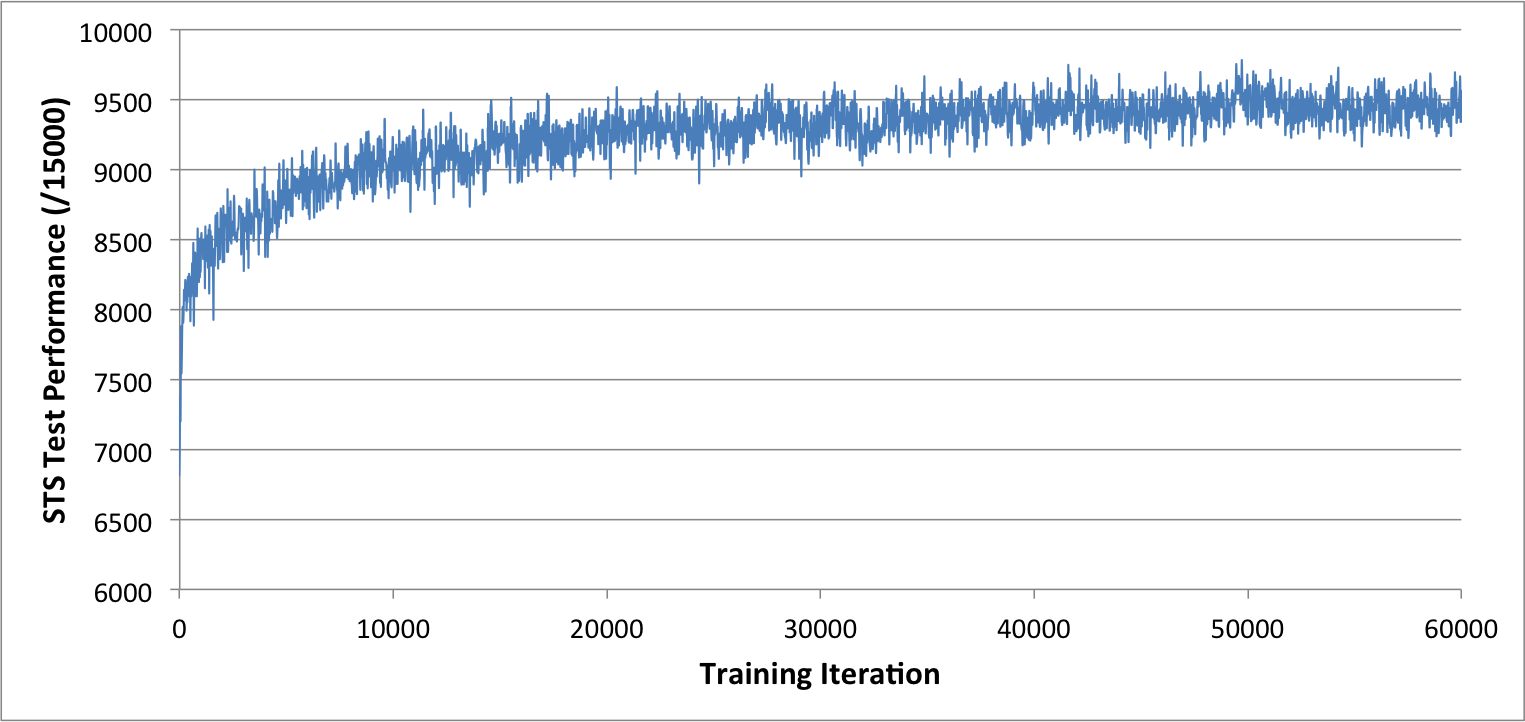}
    \caption{Training log}
    \label{fig:training_graph}
\end{figure}

Figure~\ref{fig:training_graph} shows the result of running the test periodically as training progresses. With the material-only bootstrap, it achieves a score of approximately 6000/15000. As training progresses, it gradually improved to approximately 9500/15000, with peaks above 9700/15000, proving that it has managed to gain a tremendous amount of positional understanding.

\FloatBarrier

The evaluator network converges in about 72 hours on a machine with 2x10-core Intel Xeon E5-2660 v2 CPU. Training is fully parallelised with linear speedup up to 20 threads.

For comparison, we also ran a selection of other engines on the test suite. Since we want to compare the quality of the evaluation function purely on the basis of the quality of the outputs (not efficiency), we ran the test on Giraffe with a few different time limits to see how well it can do when it is able to search as many nodes as other engines.

\begin{table}[ht]
\centering
\caption{STS Test Results (0.1s/position)}
\label{tab:sts_res}
\begin{tabular}{|l|l|l|l|}
\hline
{\bf Engine}            & {\bf Approx. Rating \cite{ccrl4040}} & {\bf Average Nodes Searched} & {\bf STS Score} \\ \hline
Giraffe (1s)            & 2400                         & 258570                       & 9641            \\ \hline
Giraffe (0.5s)          & 2400                         & 119843                       & 9211            \\ \hline
Giraffe (0.1s)          & 2400                         & 24134                        & 8526            \\ \hline
Stockfish 5             & 3387                         & 108540                       & 10505           \\ \hline
Senpai 1.0              & 3096                         & 86711                        & 9414            \\ \hline
Texel 1.04              & 2995                         & 119455                       & 8494            \\ \hline
Arasan 17.5             & 2847                         & 79442                        & 7961            \\ \hline
Scorpio 2.7.6           & 2821                         & 139143                       & 8795            \\ \hline
Crafty 24.0             & 2801                         & 296918                       & 8541            \\ \hline
GNU Chess 6 / Fruit 2.1 & 2685                         & 58552                        & 8307            \\ \hline
Sungorus 1.4            & 2309                         & 145069                       & 7729            \\ \hline
\end{tabular}
\end{table}

It is clear that Giraffe's evaluation function now has at least comparable positional understanding compared to evaluation functions of top engines in the world, which is remarkable because their evaluation functions are all carefully hand-designed behemoths with hundreds of parameters that have been tuned both manually and automatically over several years, and many of them have been worked on by human grandmasters. The test suite likely under-estimates the positional understanding of Giraffe compared to other engines, because most of the themes tested by the test suite are generally well-understood concepts in computer chess that are implemented by many engines, and since the test suite is famous, it is likely that at least some of the engines have been tuned specifically against the test suite. Since Giraffe discovered all the evaluation features through self-play, it is likely that it knows about patterns that have not yet been studied by humans, and hence not included in the test suite.

As far as we are aware, this is the first successful attempt at using machine learning to create a chess evaluation function from self-play, including automatic feature extraction (many previous attempts are weight-tuning for hand-designed features), starting from minimal hand-coded knowledge, and achieving comparable performance to state-of-the-art expert-designed evaluation functions.

\section{Probabilistic Search}
\label{sec:prob_search}

Before we introduce the work on training neural networks for move ordering and time allocation, we have to describe how Giraffe performs searches. While it is not the main topic of investigation of this project, it is a significant departure from how searches are done in conventional chess engines. While it does provide a small benefit in playing strength by itself, the main benefit comes from the fact that it provides a much more theoretically-sound foundation for us to build the machine learning system on.

\subsection{The Search Problem Redefined}
\label{sec:search_prob}

As mentioned in Section~\ref{sec:background}, there are multiple ways to define the search problem, and they are all equivalent. In any case, the goal of the search is to predict which position will be reached if both sides play a theoretically-optimal game, and a sequence of moves leading up to that position. That is because if we can have the list of moves that make up the theoretically-optimal game from the start position of the search, the first move of that list is obviously the move we should play, given the assumption that the opponent is at least as skilled as we are.

We can apply this goal recursively until we reach all possible ending positions in chess. We can then define a theoretical principal variation (PV) of any position to be the sequence of theoretically-optimal moves that will take us from the position under consideration to an end of game (checkmate or draw by stalemate/insufficient material/repetition/no progress). Some positions may have more than one theoretical PV, but they all have at least one. For the purpose of this discussion, we will assume that they all have exactly one. This assumption is unavoidable because in general, there is no way to know how many theoretical PVs there are without searching the entire search space, which is not practical in general.

Therefore, we can redefine the goal of the search to be that we want to see as much of the theoretical PV as possible. Given that in most cases we do not have time to search the entire search space and compute the actual theoretical PV, we have to decide which subset of the theoretical complete search tree to search.

The main difference between Giraffe's probabilistic search and conventional depth-limited search is in how they decide which parts of theoretical search trees to search.

\subsection{Conventional Depth-Limited Search}
\label{sec:depth_limited_search}

When von Neumann introduced the minimax algorithm in 1928, he focused on cases where the entire game tree can be explored \cite{neumann1928theorie}. While that is an important achievement, it is not applicable for games such as chess, where the game tree is too large to be completely explored. Norbert's rendition of minimax in 1948 is the first known application of incomplete minimax that uses a heuristic function at leaves of the search tree \cite{wiener1948cybernetics}. He used distance-to-root as a simple metric for determining which part of the subtree to search.

While modern chess engines have improved on that by introducing various types of extensions and reductions \cite{goetsch1990experiments} \cite{smith1994analysis} \cite{heinz2000extended}, the fundamental idea of using depth to define search trees has not changed. Virtually all existing chess engines perform depth-limited searches.

Pseudo-code for a simplified depth-limited search:
\begin{lstlisting}[caption=Depth-limited minimax, label=dl_minimax2, mathescape]
function minimax(position, depth)
{
    if depth == 0:
        return evaluate(position)
        
    bestScore = -$\infty$
                
    for each possible move mv:
        subScore = -minimax(position.apply(mv), depth - 1)
        if subScore > bestScore:
            bestScore = subScore
                
    return bestScore
}
\end{lstlisting}

\subsection{Probability-Limited Search}
\label{sec:prob_limited_search}

In Giraffe, we introduce a new way of defining search tree boundary - using a probability threshold instead of depth. Instead of searching "all nodes less than 10 moves from the root position", we search "all nodes with greater than $0.0000001$ chance of being part of the theoretical PV from the root position". 

At every position inside the search, we divide up the probability estimate for the node into probability estimates for children of the node. Probability estimates for all children of a node must add up to the probability estimate of the parent. This is because if $P(parent)$ is the probability of a parent position being part of the theoretical PV, and the parent has $n$ children, $P(child_{1}|parent)$ + $P(child_{2}|parent)$ + ... + $P(child_{n}|parent)$ = $1$, if we assume that there is exactly one theoretical PV. 

From elementary probability theory, we also know that for each child $i$, $P(child_{i} \& parent) =  P(child_{i}|parent) P(parent)$. Therefore, all we need to propagate probabilities down the search tree is a way to estimate $P(child_{i}|parent)$, for all child $i$ of any parent.

As a starting point, we assume that we do not have any information about the position or the moves, and have to assume that each child of the same parent would have the same probability.

Pseudo-code for a probability-limited search (differences highlighted):
\begin{lstlisting}[caption=Probability-limited minimax, label=pl_minimax, mathescape, escapechar=@]
function minimax(position, @\textcolor{blue}{currentProbability}@)
{
    if @\textcolor{blue}{currentProbability < ProbabilityThreshold}@:
        return evaluate(position)
        
    bestScore = -$\infty$
    
    @\textcolor{blue}{numMoves = number of legal moves from this position}@
                
    for each possible move mv:
        subScore = -minimax(position.apply(mv), @\textcolor{blue}{currentProbability / numMoves}@)
        if subScore > bestScore:
            bestScore = subScore
                
    return bestScore
}
\end{lstlisting}

\subsection{Depth-Limited Search vs Probability-Limited Search}
\label{sec:depth_vs_prob}

In most cases, probability-limited search and depth-limited search will search approximately the same sub-tree. This is because in most cases, the branching factor is approximately constant throughout the search, and node probabilities will decrease at a uniform rate, and the search will terminate at approximately uniform depth.

However, they perform differently when there are positions with subtrees with different branching factors. One extreme example is check - when a check is given, the opponent usually only has a few evasion moves available. Another example is when one branch involves exchanging high mobility pieces (eg. queens). In those instances, branches after the exchange would have much lower branching factor.

Figure~\ref{fig:dls_vs_pls} below illustrates how the search tree would look in such a situation, and which nodes a depth-limited search and a probability-limited search will choose to explore.

\FloatBarrier
\begin{figure}[ht!]
    \centering
    \includegraphics[scale=0.5]{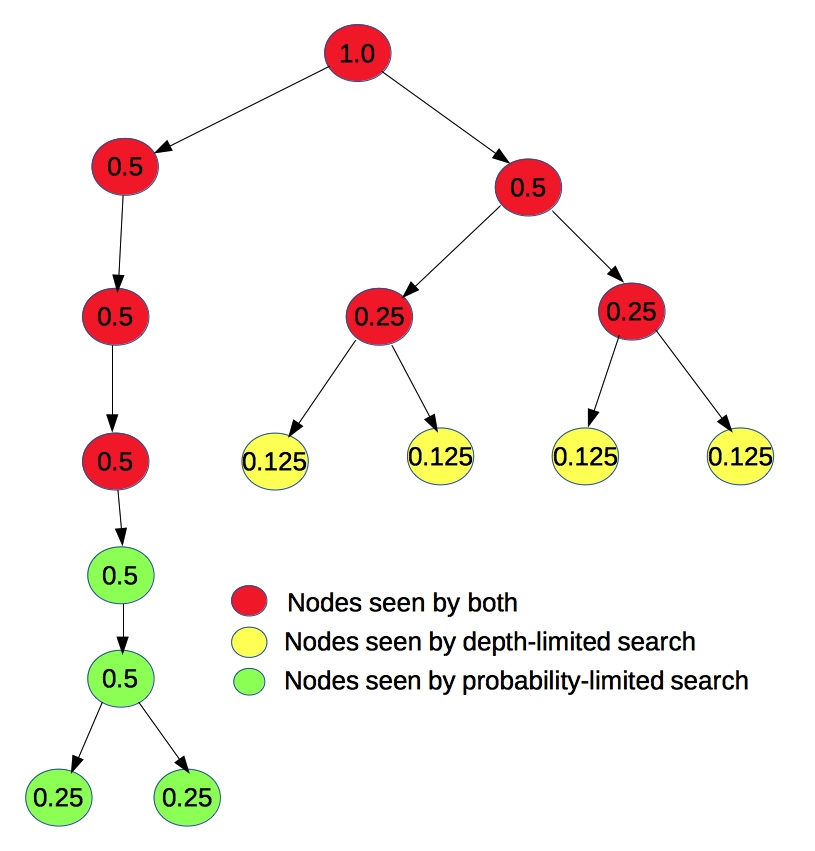}
    \caption{Example of nodes searched by depth-limited search vs probability-limited search}
    \label{fig:dls_vs_pls}
\end{figure}

Depth-limited searches will spend much more time on subtrees with high branching factor, even though there is no indication that the branch is likely going to be taken. As an extreme example, consider the situation in Figure~\ref{fig:qexchg} below, where white has the option of exchanging the queens, or not exchanging the queens. A depth-limited search would spend almost all its time searching the branch with queens still on the board, even though it could have searched much deeper in the branch with the exchange, using only a tiny fraction of the time it's spending on the no-exchange branch, and the information from searching deeper on the exchange branch is much more likely going to be useful.

A probability-limited search, on the other hand, would spend roughly equal time on both branches, and reach much higher depth on the queen-less branch. From a probabilistic perspective this behaviour makes more sense, because each node at depth 5 on the no-exchange branch is much less likely to be part of the theoretical PV than the nodes at depth 7 on the exchange branch, simply because the branching factor on the exchange branch is much lower.

\FloatBarrier
\begin{figure}[ht!]
    \centering
    \includegraphics[scale=1.0]{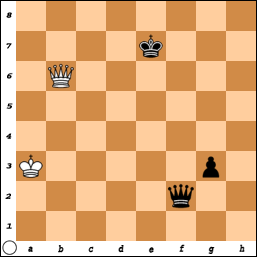}
    \caption{Example of search tree with very different branching factors in different branches}
    \label{fig:qexchg}
\end{figure}

Testing showed that a naive implementation of probability-limited search is slightly ($26 \pm 12$ rating points, corresponding to a 53\%-47\% score split) stronger than a naive implementation of depth-limited search\footnote{Please see Appendix~\ref{app:elo} for a description of the Elo rating system.}. This is a significant difference considering the fact that there should not be any difference in the vast majority of positions. No comparison has been done between an advanced implementation of probability-limited search vs depth-limited search due to time constraint.

However, most advanced search optimizations used in depth-limited searches can also be implemented for probability-limited search. Interestingly, in depth-limited searches, it has been found that extending the search in cases of checks \cite{kaindl1983searching} and situations where there is only a single reply \cite{heinz2000darkthought} is beneficial. All those techniques make a depth-limited search perform more like a probability-limited one. Probability-limited search seems to be a generalisation of those techniques.

Giraffe, as far as we know, is the first engine to implement a probability-limited search.

\section{Neural Network-Based Probability Estimation}
\label{sec:nn_prob_est}

In the previous section we explained how instead of searching all branches to equal depth, we can search all branches to an equal probability threshold instead. We previously allocated probability evenly to each children, so that:

\[
P(child_{i}|parent) = \frac{1}{n}
\]

where $n$ is the number of children for the parent.

In this section, we explore the possibility of using a neural network to estimate $P(child_{i}|parent)$. This should, in theory, give us a more selective search.

Concretely, we train a network to produce a probability $P(child_{i}|parent)$ given a parent position, as well as information about a move (that will lead to the child). Then, during the search, for each position, we would use it to evaluate each legal move, and normalize it to produce a probability distribution, which we then multiply by the probability of the parent to produce a probability for each child.

\subsection{Feature Representation}
\label{nnbpe_feature_rep}

The feature representation for this function consists of two parts - features that describe the parent position, and features that describe the move under consideration.

The part that describes the parent position is exactly the same as the feature representation we use for position evaluation, as described in Section~\ref{sec:representation}. Additional features are as follows:

\begin{itemize}
\item \textbf{Piece Type} - Which type of piece is it?
\item \textbf{From Square} - "From" square of the move
\item \textbf{To Square} - "To" square of the move
\item \textbf{Promotion Type} - For promotions, the type the pawn is promoting into
\item \textbf{Rank} - How does the move compare to other legal moves?
\end{itemize}

Most of the additional features are self-explanatory, but the Rank feature is a little curious.

The probability that a move is the best in a certain situation depends on other legal moves for that position. For example, if there is a four-way tie for most likely best moves for a position, any move that is significantly worse than the four likely will not need to be considered at all, because even if some of the four moves turn out to be poor, chances are good that at least one of them would still be better than this move. On the other hand, if there is only one better move, this move would have a significant probability of being the best move, no matter how much better the first move is, simply because the probability estimation process is not perfect.  

However, we now have a chicken-and-egg situation. How do we use the ranking of the moves to compute probabilities, when we want to use the probabilities to produce the ranking? Our solution is a two-pass process. On the first pass, all moves are evaluated as if they are the best move. We then rank the moves based on those probabilities, and then evaluate all moves again based on the ranking.

In depth-limited searches, this insight is often implemented by arbitrarily reducing search depth as the engine goes down the move list. This technique is known as late move reduction \cite{hoki2012efficiency}. Some engines (including both of the current top-2 engines Stockfish and Komodo\cite{talkchesslmp}) even go as far as not searching some moves at all, when they are at the end of long move lists. It is a crude technique that works quite well in practice. In Giraffe, we are trying to achieve the same in a more civilised way.

\subsection{Network Architecture}
\label{sec:search_network_arch}

Network architecture for the probability estimating network is very similar to the architecture for the evaluation network described in Section~\ref{sec:network_arch}. The only difference is that the output node has logistic activation instead of hyperbolic tangent activation. This shifts the output to the range (0, 1) for computational reasons (simpler calculation of gradients with cross-entropy loss).

It also has rectified linear activation for hidden nodes, and separately-connected layers for features from the three different modalities. Please see Section~\ref{sec:network_arch} for a more detailed explanation of this connectivity scheme.

\subsection{Training Positions Generation}
\label{sec:search_network_training_pos_generation}

For training the probability estimation network, we also need a corpus of unlabeled positions. However, we do not use the same corpus as the one we used for evaluation network training, because the distributions are different.

For training the evaluation network using TD-Leaf($\lambda$), we need the training positions to be representative of root positions that will appear in actual games. That means it will lack positions with heavy material imbalance for example, because those positions do not normally appear in games. However, those positions do appear very often in internal nodes of search trees. Since the probability estimation network is used on internal nodes, we want to have a training set that is distributed the same way.

To generate this training set, we perform time-limited searches on the root positions from the training set for evaluation network training (see Section~\ref{sec:training_set_gen}), and randomly sample from the positions encountered as internal nodes in those searches.

The training set consists of 5 million positions.

\subsection{Network Training}
\label{sec:search_network_training}

Unlike the case for the evaluation network, there is no need for iterative temporal-difference learning in this case, because we already have a trained evaluator, and can already find the best move for each training position by simply searching them using a static uniform-probability estimator.

The first step of the training process is to label each training position with the best move by performing a time-limited search on each one of them.

Once we have the labeled positions, we generate the training set for gradient descent by combining each position with all legal moves from that position, labeled by a binary training target specifying whether the move is the best move or not.

We then perform stochastic gradient descent with AdaDelta update rules using the cross-entropy loss function.

\subsection{Results and Evaluation}
\label{sec:prob_network_results}

The trained network turned out to be very effective. Table~\ref{tab:move_pred_res_full} below shows the predictive power of the network. 46\% of the time, the actual best move turned out to be the predicted best move. 70\% of the time, the actual best move turned out to be among the top 3 predicted moves.

\begin{table}[ht]
\centering
\caption{Move prediction results (all move types)}
\label{tab:move_pred_res_full}
\begin{tabular}{|l|l|l|}
\hline
Predicted Rank of Actual Best Move & Frequency (\%) & Cumulative Frequency (\%) \\ \hline
0                                  & 45.73          & 45.73                     \\ \hline
1                                  & 15.95          & 61.68                     \\ \hline
2                                  & 7.91           & 69.59                     \\ \hline
3                                  & 5.18           & 74.77                     \\ \hline
4                                  & 3.71           & 78.48                     \\ \hline
5                                  & 3.00           & 81.48                     \\ \hline
6                                  & 2.38           & 83.86                     \\ \hline
7                                  & 1.77           & 85.63                     \\ \hline
8                                  & 1.86           & 87.49                     \\ \hline
9                                  & 1.35           & 88.84                     \\ \hline
10                                 & 1.54           & 90.38                     \\ \hline
Below 10th                         & 9.62           & 100.00                     \\ \hline
\end{tabular}
\end{table}

Table~\ref{tab:move_pred_res_quiet} shows the result on a different testing set, where all positions where the best move is a winning capture (easy positions) have been filtered out. Even among these difficult quiet positions, the correct best move is predicted 28\% of the time, and the best move is among the top 3 predicted moves 55\% of the time.

\begin{table}[ht]
\centering
\caption{Move prediction results (quiet moves)}
\label{tab:move_pred_res_quiet}
\begin{tabular}{|l|l|l|}
\hline
Predicted Rank of Actual Best Move & Frequency (\%) & Cumulative Frequency (\%) \\ \hline
0                                  & 28.37          & 28.37                     \\ \hline
1                                  & 15.97          & 44.35                     \\ \hline
2                                  & 11.03           & 55.38                     \\ \hline
3                                  & 7.99           & 63.37                     \\ \hline
4                                  & 5.94           & 69.31                     \\ \hline
5                                  & 4.01           & 73.32                     \\ \hline
6                                  & 3.59           & 76.91                     \\ \hline
7                                  & 3.13           & 80.05                     \\ \hline
8                                  & 2.86           & 82.90                     \\ \hline
9                                  & 2.45           & 85.35                     \\ \hline
10                                 & 1.99           & 87.34                     \\ \hline
Below 10th                         & 12.66          & 100.00                     \\ \hline
\end{tabular}
\end{table}

In terms of actual playing strength, with the negative effect on search speed excluded, Giraffe is much stronger with the neural network probability estimator. After 3000 games, the difference is $48 \pm 12$ Elo points, which corresponds to a score split of 57\% vs 43\%\footnote{Please see Appendix~\ref{app:elo} for a description of the Elo rating system.}.

However, the neural network is too slow to be used everywhere in a search tree. When used near the leaves, it can spend more time allocating probability to the nodes than actually searching the nodes. Therefore, it is only used on nodes with more than about 100 nodes to search below them. In this case, the improvement in strength is approximately $20 \pm 12$ Elo points.

\section{Conclusion}
\label{sec:conclusion}

In this project, we investigated the use of deep reinforcement learning with automatic feature extraction in the game of chess. The results show that the learned system performs at least comparably to the best expert-designed counterparts in existence today, many of which have been fine tuned over the course of decades.

The beauty of this approach is in its generality. While it was not explored in this project due to time constraint, it is likely that this approach can easily be ported to other zero-sum turn-based board games, and achieve state-of-art performance quickly, especially in games where there has not been decades of intense research into creating a strong AI player.

In addition to the machine learning aspects of the project, we introduced and tested an alternative variant of the decades-old minimax algorithm, where we apply probability boundaries instead of depth boundaries to limit the search tree. We showed that this approach is at least comparable and quite possibly superior to the approach that has been in use for the past half century. We also showed that this formulation of minimax works especially well with our probability-based machine learning approach.

Efficiency is always a major consideration when switching from an expert system to a machine learning approach, since expert systems are usually more efficient to evaluate than generic models. This is especially important for applications like a chess engine, where being able to search nodes quickly is strongly correlated with playing strength. Some earlier attempts at applying neural network to chess have been thwarted by large performance penalties. For example, Thrun reported a difference of 2 orders of magnitude between his implementation of a neural-network based chess engine and a state of art chess engine of that time \cite{thrun1995learning}. Giraffe's optimized implementation of neural network, when combined with the much higher vector arithmetics throughput of modern processors and effective caching, allows it to search at a speed that is less than 1 order of magnitude slower than the best modern chess engines, thus making it quite competitive against many chess engines in gameplay without need for time handicap.

With all our enhancements, Giraffe is able to play at the level of an FIDE International Master on a modern mainstream PC. While that is still a long way away from the top engines today that play at super-Grandmaster levels, it is able to defeat many lower-tier engines, most of which search an order of magnitude faster. One of the original goals of the project is to create a chess engine that is less reliant on brute-force than its contemporaries, and that goal has certainly been achieved. Unlike most chess engines in existence today, Giraffe derives its playing strength not from being able to see very far ahead, but from being able to evaluate tricky positions accurately, and understanding complicated positional concepts that are intuitive to humans, but have been elusive to chess engines for a long time. This is especially important in the opening and end game phases, where it plays exceptionally well.

\subsection{Future Work}

Many interesting ideas and potential improvements for Giraffe were not implemented due to time constraint. This section lists some of those ideas.

\subsubsection{Neural Network-Based Time Management}

Time management is an important part of chess playing. Different tournaments use different time controls (for example, common time controls for online play are 5 minutes per side for the entire game, or 2 minutes per side with 12 seconds added after every move). However, they all require players to divide up the allocated time and decide how much time to use for every move. There are many considerations - complexity of the position, expected remaining length of the game, and confidence in the currently chosen move (while the thinking is ongoing).

Giraffe and virtually all other chess engines today use simple heuristics to decide how much time to spend per move. It may be beneficial to use a neural network to make these decisions instead.

\subsubsection{More Fine-Grained Caching, and Model Compression}

One of the major weaknesses of Giraffe is the search speed. Although it is already competitive against many engines at its very low speed, it would be much stronger if it can close the speed gap.

Giraffe already caches neural network outputs. However, hit rate of the cache is low because exact same positions are rarely evaluated twice. However, most positions encountered in a search tree will be similar, and with clever designs of network connectivity, it may be possible to cache intermediate activations within networks.

Some recent discoveries in neural network training by Ba and Caruana suggest that it may be possible to compress large neural networks by using them as mentors to train smaller networks \cite{ba2014deep}. This is another potential way to speed up Giraffe.

\subsubsection{Evaluation Error Estimation}

When the static evaluation function is called to evaluate a position, it only needs to return an accurate evaluation if it is within a set of bounds supplied by the caller. Due to the way $\alpha$-$\beta$ pruning works, if the true score is going to be outside the bounds, the caller only needs to be informed that it is going to be outside the bounds.

Therefore, it may be beneficial to be able to quickly estimate the upper- and lower-bound of a score of a position, since if the possible score range is entirely out of the $\alpha$-$\beta$ window, there is no need to perform the expensive full evaluation.

One possible way to perform bound estimation is by training a network with two outputs - the two bounds. The two outputs can both be trained to predict the evaluator output, but with different error functions. For example, asymmetrical cost functions that penalizes positive errors much more than negative errors can be used to train the output for lower-bound, and vice versa.

\subsubsection{Similarity Pruning}

Although Giraffe already has much smaller search trees than other engines of similar strength, they are still many orders of magnitudes larger than what humans can achieve. 
One of the causes for the difference is in how accurately positions are evaluated. When positions are not evaluated accurately, deeper and wider searches are required to compensate. Closing this gap has been the primary focus of this project.

Another reason for humans' high search efficiency is the concept of position similarity. Humans can often decide that some moves are effectively equivalent in some situations, and avoid searching all of them. This dramatically reduces the average branching factor of search trees.

One possible way to implement this using machine learning is to use a neural network that takes positions as inputs, and outputs sequences of numbers that can work as "signatures" for each position. Unsupervised learning (for example, clustering) can then be used on the signatures to gauge degrees of similarity between positions. However, it is unclear how such networks can be trained.

Being able to accurately predict equivalence between positions and moves, whether using machine learning or other techniques (such as inductive reasoning), is likely going to lead to another major milestone in achieving more efficient searches.

\newpage
\begin{appendices}
\section{The Elo Rating System}
\label{app:elo}

The Elo rating system is a mathematical model that assigns ratings to players based on game results. The system ensures that a stronger player would receive a higher rating than a weaker player. Ratings are only comparable within the same pool of players. A rating difference of 200 corresponds to a score split of about 75\%/25\%, with the stronger player expecting to receive 75\% of the points.

In this paper, unless otherwise specified, all ratings are on the FIDE scale.

On the FIDE scale, a beginner who knows the rules of the game would have a rating of about 1000. A serious low level tournament player would have a rating of about 1800. Magnus Carlsen, the reigning World Chess Champion, has a rating of 2853 \cite{fide_rating} (September 2015 list). The best chess engines are estimated to have ratings above 3200. However, there are not enough games between computers and humans with official FIDE ratings to establish an accurate relationship between human and computer rating pools.

FIDE awards titles to players who achieve certain ratings (with other conditions, and with exceptions) \cite{fide_rating} -

2500: Grandmaster (approx. top 0.68\% of tournament chess players with an official rating)

2400: International Master (approx. top 2.2\%)

2300: FIDE Master (approx. top 5.3\%)

2200: Candidate Master

\end{appendices}

\newpage 

\bibliographystyle{unsrtnat}
	\bibliography{references} 

\end{document}